# SoftPool++: An Encoder–Decoder Network for Point Cloud Completion

Yida Wang[1] · David Joseph Tan[2] · Nassir Navab[1] · Federico Tombari[1,2]

## Abstract
We propose a novel convolutional operator for the task of point cloud completion. One striking characteristic of our approach is that, conversely to related work it does not require any max-pooling or voxelization operation. Instead, the proposed operator used to learn the point cloud embedding in the encoder extracts permutation-invariant features from the point cloud via a *soft-pooling* of feature activations, which are able to preserve fine-grained geometric details. These features are then passed on to a decoder architecture. Due to the compression in the encoder, a typical limitation of this type of architectures is that they tend to lose parts of the input shape structure. We propose to overcome this limitation by using skip connections specifically devised for point clouds, where links between corresponding layers in the encoder and the decoder are established. As part of these connections, we introduce a transformation matrix that projects the features from the encoder to the decoder and vice-versa. The quantitative and qualitative results on the task of object completion from partial scans on the ShapeNet dataset show that incorporating our approach achieves state-of-the-art performance in shape completion both at low and high resolutions.

## 1 Introduction

Several data representations exist for 3D shapes. One common choice is the use of spatially discretized representations such as volumetric data (Yang et al. 2017; Wang et al. 2019b; Yang et al. 2018a). Alternative popular choices are implicit descriptions (Park et al. 2018; Chibane et al. 2020) as well as sparse 3D coordinate-based representations such as point clouds (Yang et al. 2018b; Xie et al. 2020b; Yuan et al. 2018) and 3D meshes (Groueix et al. 2018). Among this latter category of 3D data formats, point clouds are arguably the simplest, since they store 3D coordinates without any additional topological information such as faces or edges associated to the vertices. Hence, investigating how to process and learn 3D shape geometry based on these simple, yet effective representations is currently a hot research topic. This has recently motivated several tasks in 3D computer vision such as estimating point cloud deformation (Yang et al. 2018b; Yuan et al. 2018), registration (Aoki et al. 2019; Park et al. 2017), completion (Wang et al. 2020b; Groueix et al. 2018; Yuan et al. 2018), segmentation (Qi et al. 2017a; Lei et al. 2020; Xu et al. 2020) and 3D object detection (Shi et al. 2020; Qi et al. 2019).

This paper focuses on the point cloud completion task. The goal is to fill out occluded parts of the input 3D geometry represented by a partial scan, in a way that is coherent with the global shape while preserving fine local surface details. This is a useful task for many real world applications since occluded regions are normally present as part of most 3D data capture processes within, e.g., SLAM or multi-view reconstruction pipelines. State-of-the-art approaches targeting this task are based on neural networks and mostly rely on learning how to deform a set of 2D grids at different scales into 3D points, based on global shape descriptors typically represented by PointNet (Qi et al. 2017a) features. Examples of these approaches are FoldingNet (Yang et al. 2018b), AtlasNet (Groueix et al. 2018) and PCN (Yuan et al. 2018).

To overcome the aforementioned problem related to information loss due to feature compression at the level of the encoder–decoder bottleneck, GRNet (Xie et al. 2020b) suggests to preserve fine geometry details by discretizing the features via volumetric feature maps used at the different layers of the encoder. It also suggests using volumetric U-Net (Yang et al. 2018a) to build skip connections between the encoder and the decoder, eventually merging the obtained features with the input point cloud. The idea of leveraging skip connections among different layers of an encoder–decoder model follows the successful paradigm already exploited for volumetric shape completion, in particular 3D-RecGAN (Yang et al. 2017) and ForkNet (Wang et al. 2019b). While effective, converting sparse point cloud features into volumetric maps brings in all the disadvantages of discretized 3D data representations with respect to point clouds, in particular the loss of fine shape details, the inability to flexibly deal with local point density variations, as well as the unpractical trade-off between 3D resolution and memory occupancy.

Recently, we have demonstrated how, by means of sorting features based on their activations rather than applying max pooling, we can build up point clouds embeddings that store more informative features for a point cloud with respect to PointNet. This feature-learning approach, named SoftPool (Wang et al. 2020b), obtained state-of-the-art results for different point cloud-related tasks, such as completion and classification. In this work, we build up on our previous work (Wang et al. 2020b) to propose a more complete end-to-end framework. Our contributions are two-folds and are listed as follows:

1. We generalize our feature extraction technique into a module called SoftPool++. This module introduces *truncated softpool features* aimed to decrease the memory requirements of the original method during training, making it compatible with off-the-shelf GPUs. Notably, a disadvantage of the SoftPool features (Wang et al. 2020b) is that each point is processed independently from the rest. Due to this, the proposed module further processes the truncated softpool features with regional convolutions in order to recognize the relationships between the feature points. In contrast to Wang et al. (2020b) that applies their feature once, this module can be applied multiple times as demonstrated in our architecture, which uses it across multiple layers.

2. We propose a novel encoder–decoder architecture characterized by the use of point-wise skip connections. By connecting corresponding layers between encoder and decoder, this has the advantage of preserving fine geometric details from the given partial input cloud. This is to the best of our knowledge the first approach using skip connections for unorganized sets of 3D feature maps, relaxing the need of spatial discretization as deployed in Xie et al. (2020b), with benefits in terms of completion accuracy and memory occupancy. In addition, we also adapt the discriminator from TreeGAN (Shu et al. 2019) for the shape completion problem to further improve our model.

Our method is evaluated on ShapeNet (Chang et al. 2015) for the task of shape completion and on ModelNet (Zhirong et al. 2015) and PartNet (Mo et al. 2019) for the task of classification. Figure 1 illustrates a teaser of the shape completion results. It compares the architectures that are built on PointNet (Qi et al. 2017a) and SoftPool (Wang et al. 2020b) features. Visually, we show the advantage of the reconstructions that rely on SoftPool features as they are remarkably more similar to the ground truth. Moreover, the figure also highlights the improvements of SoftPool++ with respect to our previous approach (Wang et al. 2020b).

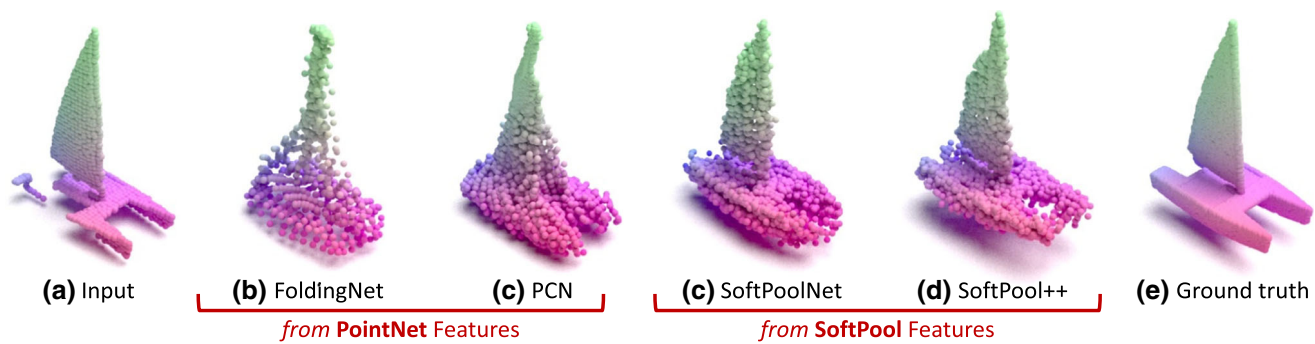

**(a)** Input  **(b)** FoldingNet  **(c)** PCN  **(c)** SoftPoolNet  **(d)** SoftPool++  **(e)** Ground truth

*from* **PointNet** Features  *from* **SoftPool** Features

**Fig. 1** Object completion results of the PointNet features such as FoldingNet (Yang et al. 2018b) and PCN (Yuan et al. 2018); and, the SoftPool features such as SoftPoolNet (Wang et al. 2020b) and the proposed SoftPool++

## 2 Related Work

Based on the focus of our contributions, we browse through the relevant methods in 3D object completion from partial scans and the use of skip connections with 3D data.

### 2.1 3D Object Completion

Inspired by the way humans perceive the 3D world from 2D projections, 3D-R2N2 (Choy et al. 2016) builds recurrent neural networks (RNNs) to fuse multiple feature maps extracted from input RGB images sequentially to recover the 3D geometries. To further improve the reconstruction, a coarse-to-fine 3D decoder was presented in Pix2Vox (Xie et al. 2019) as well as the residual refiner in Pix2Vox++ (Xie et al. 2020a). Due to the recent popularity of the attention mechanisms, AttSets (Yang et al. 2020) proposed to build attention layers to correlate the image features from different views. In contrast, our 3D reconstruction in this paper focuses on only a single depth image.

Taking a depth image of an object from an arbitrary camera pose, the objective of 3D object completion is to complete its missing structure and build its full reconstruction. Focusing on learning-based completion, most related work can be categorized depending on the input data they process—voxelized grid or point cloud. Interestingly, a notable work from OcCo (Wang et al. 2021a) demonstrates that the weights trained for completion are also valuable for other tasks like segmentation and classification.

**Voxelized Grid** Due to the popularity of 2D convolution operations in CNNs (Azad et al. 2019; Kirillov et al. 2020; Yang et al. 2020) for RGB images, its straightforward extension to 3D convolutions on volumetric data also rose to fame. 3D-EPN (Dai et al. 2017) and 3D-RecGAN (Yang et al. 2017) are the first works on this topic, where they extended the typical encoder–decoder architecture (Noh et al. 2015) to 3D. Adopting a similar architecture, 3D-RecGAN++ (Yang et al. 2018a) and ForkNet (Wang et al. 2019b) utilize adversarial training with 3D discriminator to improve the reconstruction.

The main advantage of volumetric completion is the structure of its data such that deep learning methods developed for RGB images can be extended to 3D. However, this advantage is also its limitation. The fixed local resolution makes it hard to reconstruct the object's finer details without consuming a huge amount of memory.

**Point Cloud** Having the inverse problem, point clouds have the potential to reconstruct the object at a higher resolutions but exhibited so far a limited application in deep learning due to its unstructured data. Note that, unlike RGB images or voxel maps, point clouds do not have a particular order, and the number of points varies as we change the camera pose or the object.

Targeted to solve the unordered structure of point clouds, PointNet (Qi et al. 2017a) proposes to implement max-pooling in order to achieve a permutation invariant latent feature. Based on this one dimensional feature, FoldingNet (Yang et al. 2018b) proposes an object completion solution that deforms a 2D rectangular grids by multi-layer perceptron (MLP). By increasing the number of 2D rectangular grids, AtlasNet (Groueix et al. 2018) and PCN (Yuan et al. 2018) added more complexity as well as details into the reconstruction. MSN (Liu et al. 2020) then further improves the completion by adding restrictions to separate different patches apart from each other. Moreover, Cycle4Completion (Wen et al. 2021) is also based on PointNet features but solves the problem by training with an unsupervised cycle transformation. Moving away from the global feature representation, PointNet++ (Qi et al. 2017b) samples the local subset of points with farthest point sampling (FPS) then feeds it into PointNet (Qi et al. 2017a). Based on this feature, PMPNet (Wen et al. 2020b) completes the entire object gradually from the observed regions to the nearest occluded regions. SnowflakeNet (Xiang et al. 2021) also uses the PointNet++ features to split points in the coarsely reconstructed object to execute the completion progressively. In addition, building a similar feature as PointNet, ME-PCN (Gong et al. 2021) takes both the occupied and the empty regions on the depth image as input for 3D completion, showing the advantage of masking the empty regions in completion.

Unlike the methods which are dependent on a vectorized global feature to solve the permutation invariant problem, RFNet (Huang et al. 2021) and PointTr (Yu et al. 2021) produce several global features in their encoder. On one hand, RFNet (Huang et al. 2021) uses their features to complete the object in an recurrent way by concatenating the incomplete input and the predicted points level by level. On the other, PointTr (Yu et al. 2021) relies on transformers to produce a set of queries directly from the observed points with the help of positional coding. In effect, PointTr (Yu et al. 2021) does not need to compress the input into a single vector.

The recent work from PVD (Zhou et al. 2013), GRNet (Xie et al. 2020b) and VE-PCN (Wang et al. 2021b) leverage both the point cloud and the voxel grid representations. Unlike most works that rely on Chamfer distance to optimize the model, PVD (Zhou et al. 2013) uses a simple Euclidean loss to optimize the shape generation model from the voxelized point cloud representation. GRNet (Xie et al. 2020b) first voxelizes the point cloud, processes the voxel grid with deep learning and converts the results back to point cloud. While this solves the unorganized structure of the point clouds, its discretization removes its advantage on reconstructing in higher resolutions. VE-PCN (Wang et al. 2021b) improves the completion by supplementing the features of the decoder in the volumetric completion with the edges. This method then converts the voxels to point clouds by Adaptive Instance Normalization (Lim et al. 2019).

Another solution is presented in our previous work Soft-PoolNet (Wang et al. 2020b) that builds local groups of features by sorting them into a feature map. 2D convolutions are then applied to the feature map. Consequently, this approach is able to deal with unorganized point clouds and achieve reconstruction results at high resolution. We build upon SoftPoolNet (Wang et al. 2020b) and generalize the feature extraction into a module which we call SoftPool++. This then allows us to connect multiple modules in an encoder–decoder architecture. As a consequence, we achieve better quantitive and qualitative results.

## 2.2 Skip Connections in 3D

Skip connections were initially proposed for image processing (Mazaheri et al. 2019; Kim et al. 2016; Gao et al. 2019; Azad et al. 2019) then later adapted for 3D volumetric reconstruction (Yang et al. 2017, 2018a; Wang et al. 2019b). Given a point cloud as input, the methods like GRNet (Xie et al. 2020b) and InterpConv (Mao et al. 2019) require to convert the input point cloud to voxel grids.

Aiming at alleviating this limitation on point clouds, the work from Std (Yang et al. 2019) bypasses the encoder features into decoder point-by-point while GACNet (Wang et al. 2019a) constructs a graph from the points then constructs the skip connection with the graph. The problem of these point-wise skip connections is that new points cannot be introduced in the decoder. To solve this, SA-Net (Wen et al. 2020a) groups PointNet++ (Qi et al. 2017b) features in different resolutions with KNN. The skip connection from the encoder then matches the resolution of the decoder.

Contrary to these methods, in the context of object completion, the objective of our skip connection is compensate for the lost data in the encoder and bypass the observed geometry to the decoder. We also introduce the concept of feature transformation to compensate for the difference between the features from the encoder and decoder. Later in our evaluation, we found that the skip connection is a crucial step to achieve higher accuracy. Moreover, the SoftPool++ features also contribute to make our skip connection simpler. Since it is an organized feature, we avoid the time-consuming KNN, which significantly decreases our inference time.

## 3 Feature Extraction

Given the partial scan of an object, the input to our network is a point cloud with $N_{in}$ points written in matrix form as $\mathbf{P}_{in} = [\mathbf{x}_i]_{i=1}^{N_{in}}$, where each point is represented as the 3D coordinates $\mathbf{x}_i = [x_i, y_i, z_i]$.

On one hand, the first objective of this section is to build a feature descriptor from the unorganized point cloud such that the feature remains the same for any permutation of the point cloud in $\mathbf{P}_{in}$. On the other hand, the second objective is to generalize this process into a feature extraction module that takes an arbitrary input $\mathbf{P}_{in}$. In this way, the proposed module can be implemented at multiple instances in our architecture.

### 3.1 SoftPool Feature

From the point cloud vector, we then convert each point into a feature vector $\mathbf{f}_i$ with $N_f$ elements by projecting every point with a point-wise multi-layer perceptron (Qi et al. 2017a) with its parameters assembled in $\mathbf{W}_{MLP}$. Thus, we define the $N_{in} \times N_f$ feature matrix as $\mathbf{F} = [\mathbf{f}_i]_{i=1}^{N_{in}}$. Note that we applied a softmax function to the output neuron of the perceptron so that the elements in $\mathbf{f}_i$ range between 0 and 1.

Throughout this section, we refer to the *toy example* in Fig. 2 to visualize the various steps. This example assumes that there are only five points in the point cloud such that $N_{in} = 5$ as shown in Fig. 2a.

One of the main challenges in processing a point cloud is its unstructured arrangement. If we look at Fig. 2a, changing the order of the points in $\mathbf{P}_{in}$ reorganizes the rows of the feature map $F$. There is consequently no guarantee that the feature map remains constant for the same set of points. To solve this problem, we propose to organize the feature vectors in $\mathbf{F}$ so that their $k$-th elements are sorted in a descending order, which is denoted as $\mathbf{F}'_k$. Note that $k$ should not be larger than $N_f$. This is demonstrated in Fig. 2a where we arrange the five feature vectors from $\mathbf{F} = [\mathbf{f}_i]_{i=1}^{5}$ to $\mathbf{F}'_k = [\mathbf{f}_i]_{i=\{3,5,1,2,4\}}$ by comparing the $k$-th element of each vector.

The features in SoftPoolNet (Wang et al. 2020b) repeat this process for all of the $N_f$ elements in $\mathbf{f}_i$. Altogether, the feature is a 3D tensor with the dimension of $N_{in} \times N_f \times N_f$ denoted as $\mathbf{F}' = [\mathbf{F}'_1, \mathbf{F}'_2, \ldots \mathbf{F}'_{N_f}]$ in Fig. 2b. Finally, we assemble the SoftPool features $\mathbf{F}^*$ by taking the $N_r$ rows with the highest activations of all $\mathbf{F}'_i$ in $\mathbf{F}'$. Since each row in $\mathbf{F}'_i$ is equivalent to a point, we can then interpret the $N_r$ rows of $\mathbf{F}'_i$ as one region in the point cloud, summing up to all $N_f$ regions in $\mathbf{F}^*$.

Although both PointNet (Qi et al. 2017a) and SoftPoolNet (Wang et al. 2020b) utilize MLP in their architecture, they have significant differences on handling the results thereof. Compared to the max-pooling operation in PointNet (Qi et al. 2017a), the motivation of the SoftPool feature is to capture a larger amount of information and to further process it with regional convolution operations, as explained later in Sect. 4.

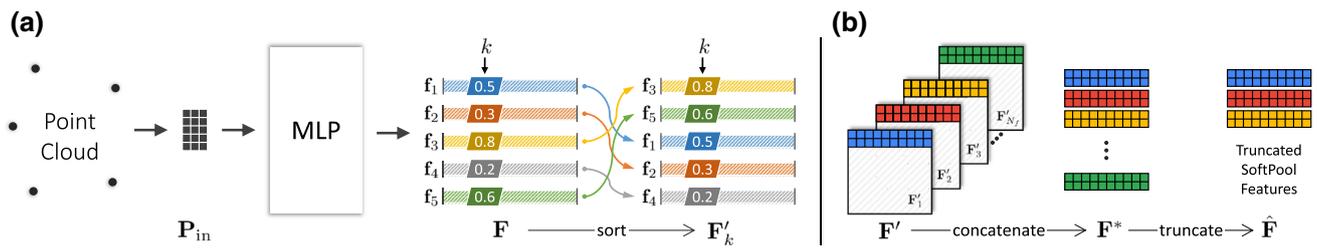

**Fig. 2** Toy examples of the truncated SoftPool feature. Given 5 points in (**a**), they go through Multi-layer Perceptron (MLP) to produce **F**. At the $k$-th element, the vectors are sorted to build $\mathbf{F}'_k$ and consequently $\mathbf{F}'$. In (**b**), we concatenation of the first $N_r$ rows of $\mathbf{F}'_k$ to construct the 3D tensor $\mathbf{F}^*$ which corresponds to the regions with high activations then truncated to assemble $\hat{\mathbf{F}}$

## 3.2 Generalizing and Truncating the SoftPool Feature

In practice, we noticed that we can generalize the SoftPool feature formulation to an arbitrary input feature $\mathbf{P}_{in}$—thus, alleviating the definition of points—to produce the Soft-Pool features $\mathbf{F}'$. From this perspective, we can construct an architecture with a series of SoftPool feature extractions. Therefore, we take the point cloud as the input to the architecture and extract the first SoftPool features. Then, after processing the first features, we can then extract the second features from them and so on. This is discussed later in Sect. 4 with an encoder–decoder architecture.

However, the drawback of such architecture is the size of the SoftPool features. With a dimension of $N_r \times N_f \times N_f$, the memory footprint increases with the size of the feature but we are constrained by the memory size of our off-the-shelf GPU. Notably, in Wang et al. (2020b), they set the feature dimension $N_f$ to a small value of 8. In this work, since we are interested in building a series of SoftPool features in an encoder–decoder architecture, $N_f$ increases up to 256 in the latent space.

Hence, we propose to further truncate the SoftPool features to $N_r \times N_f \times N_s$, where the third dimension takes the first $N_s$ matrices in $\mathbf{F}^*$ as illustrated in Fig. 2b. To distinguish from Wang et al. (2020b), we refer this as the *Truncated SoftPool feature*, denoted as $\hat{\mathbf{F}}$ in Fig. 2b.

## 3.3 Regional Convolutions

Considering that each point in the cloud independently goes through MLP while the operations thereafter to produce the truncated SoftPool features rely on sorting, each row of our feature remains independent from each other. However, in contrast to max-pooling which produces a vector, our feature is a 3D tensor which can undergo convolutional operations.

Instead of applying the same kernel to all regions as Wang et al. (2020b), we generalize the regional convolutions and impose distinct kernels for each region. We first split $\hat{\mathbf{F}} = [\hat{\mathbf{F}}_r]_{r=1}^{N_s}$ into separate regions $\hat{\mathbf{F}}_r$ and correspondingly apply a set of kernels $\mathbf{W}_{conv} = \{\mathbf{W}_r\}_{r=1}^{N_s}$. Assigning the concatenated output tensor as $\mathbf{F}_{out} = [\mathbf{P}_r]_{r=1}^{N_s}$, we can formally describe this operation as

$$\mathbf{P}_r(i,j) = \sum_{l=1}^{N_f} \sum_{k=1}^{N_k} \hat{\mathbf{F}}_r(i+k,l) \mathbf{W}_r(j,k,l) \quad (1)$$

for the $r$-th region.

The dimension of each kernel is $N_k \times N_f \times N_{out}$, where $N_k$ indicates the number of neighbors to consider and $N_{out}$ is the desired size of the output $\mathbf{P}_r$. Note that the kernels convolves on the entire width of $\hat{\mathbf{F}}_r$, i.e.corresponding to its width $N_f$. This implies that we only pad on the vertical axis. Similar to other convolutional operators, the stride $s$ distinguishes between a convolutional and deconvolutional operation. If the stride is greater than 1, $\hat{\mathbf{F}}_r$ is downsampled, while it is upsampled if the stride is less than 1.

## 3.4 SoftPool++ Module

Now, we have all the components to build the feature extraction module as shown in Fig. 3, which we call SoftPool++. Since $\mathbf{P}_{in}$ is defined as the input point cloud, we generalize the input of the module as $\mathbf{F}_{in}$ where we set $\mathbf{F}_{in} = \mathbf{P}_{in}$ in the first layer. Hence, the input matrix $\mathbf{F}_{in}$ goes through a 3-layer perceptron then builds the truncated SoftPool features. Thereafter, we perform regional convolution and reshape the results by squeezing the third dimension to finally acquire our output feature matrix $\mathbf{F}_{out}$.

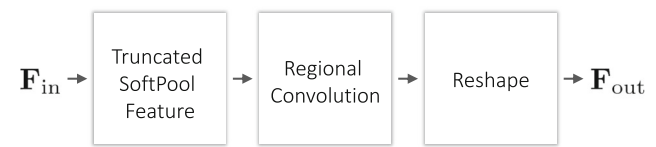

**Fig. 3** Overview of the feature extraction module called SoftPool++

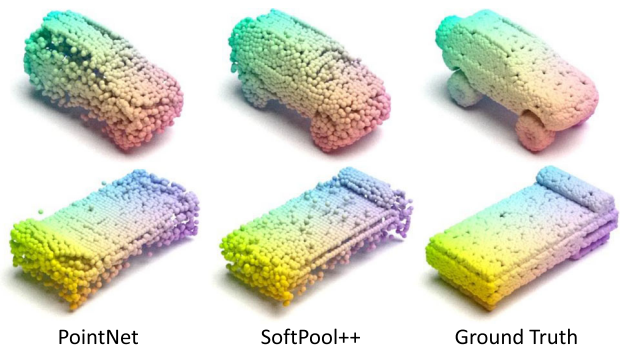

**Fig. 4** Object completion results with MSN (Liu et al. 2020) while using PointNet (Qi et al. 2017a) features and SoftPool++ features on its encoder

When constructing our architecture in Sect. 4, the encoder and decoder are distinguished primarily on the stride $s$. In this paper, we show the versatility of this novel module to act as an encoder and decoder as well as to refine a coarse point cloud with more elaborate details.

The differences between decoding from PointNet features and SoftPool++ features are evident in Fig. 4, where we replace the PointNet feature in MSN (Liu et al. 2020) with a SoftPool++ feature with the same size of 1024. By replacing the PointNet (Qi et al. 2017a) encoder in MSN (Liu et al. 2020) with our SoftPoolNet++ encoder, we show that the SoftPool++ feature supplements the MSN's decoder where all the wheels are clearly separated from the body of the SUV, while the original PointNet feature in MSN follow the more generic structure of a vehicle with tiny gaps between wheel and body. This proves that *SoftPool++* makes our decoder able to take all observable geometries into account to complete the shape, while the max-pooled PointNet feature cannot deal with geometric structures which are rarely or not at all seen in the training data.

## 4 Network Architecture

The volumetric U-Net (Çiçek et al. 2016; Yang et al. 2018a) in 3D-RecGAN (Yang et al. 2017) and GRNet (Xie et al. 2020b) has shown significant improvements in object completion as it injects more data from the encoder to the decoder in order to supplement the compressed latent feature. Without the skip connection in U-Net, we end up losing most of the input data as it goes through the encoder. Consequently, the decoder starts hallucinating the overall structure without being faithful to the given information. Inspired by this idea, we introduce a novel U-Net connection that directly takes the point cloud as input, i.e.without the need of voxelization at any stage of the network.

Our network architecture is composed of an encoder–decoder structure with a skip connection as shown in Fig. 5. Such connection between encoder and decoder makes the completion more likely to preserve input geometries. The encoder is composed of consecutive feature extraction modules from Sect. 3.4 to downsample the input to the latent feature while the decoder is composed of the similar feature extraction modules to upsample to the output. As discussed in Sect. 3.4, the stride $s$ is a significant parameter to distinguish the two layers. Table 1 lists the values of all the parameters for the module in the convolution and deconvolution layers.

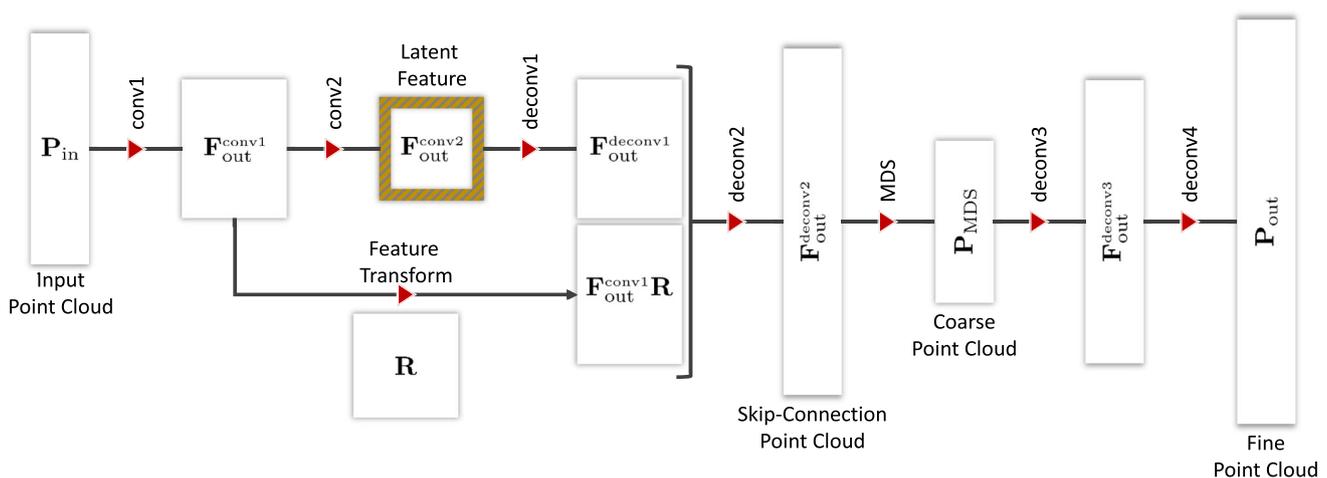

**Fig. 5** Overview of our object completion architecture where the parameters for the convolution and deconvolution operations based on the feature extraction module are listed in Table 1. Note that, in our evaluation, we compare three point cloud results from the decoder: (1) the skip-connection output; (2) the coarse output; and, (3) the fine-grained output which is the final reconstruction

**Table 1** Dimensions and parameters on each feature extraction module in our architecture

| Feature extraction module | Conv1 | Conv2 | Deconv1 | Deconv2 | MDS | Deconv3 | Deconv4 |
|---|---|---|---|---|---|---|---|
| Input variable | $\mathbf{P}_{\text{in}}$ | $\mathbf{F}_{\text{out}}^{\text{conv1}}$ | $\mathbf{F}_{\text{out}}^{\text{conv2}}$ | $[\mathbf{F}_{\text{out}}^{\text{deconv1}}, \mathbf{F}_{\text{out}}^{\text{conv1}}\mathbf{R}]$ | $\mathbf{F}_{\text{out}}^{\text{deconv2}}$ | $\mathbf{P}_{\text{MDS}}$ | $\mathbf{F}_{\text{out}}^{\text{deconv3}}$ |
| Number of rows | $N_{\text{in}}$ | $N_{\text{in}}/8$ | 256 | $512 + N_{\text{in}}/8$ | $1024 + N_{\text{in}}/4$ | 2048 | 4096 |
| Number of columns | 3 | 256 | 256 | 256 | 3 | 3 | 3 |
| Stride ($s$) | 8 | 2 | 1/2 | 1/2 | – | 1/2 | 1/4 |
| Feature dimension ($N_f$) | 256 | 256 | 256 | 256 | 3 | 256 | 256 |
| Number of rows in the region ($N_r$) | $N_{\text{in}}$ | 32 | $N_{\text{in}}$ | $N_{\text{in}}$ | – | $N_{\text{in}}$ | $N_{\text{in}}$ |
| Truncation size ($N_s$) | 1 | 8 | 1 | 1 | – | 1 | 1 |
| Kernel size ($N_k$) | 8 | 8 | 4 | 4 | – | 4 | 4 |
| Dimension of output feature ($N_{\text{out}}$) | 256 | 256 | 256 | 3 | 3 | 3 | 3 |
| Output variable | $\mathbf{F}_{\text{out}}^{\text{conv1}}$ | $\mathbf{F}_{\text{out}}^{\text{conv2}}$ | $\mathbf{F}_{\text{out}}^{\text{deconv1}}$ | $\mathbf{F}_{\text{out}}^{\text{deconv2}}$ | $\mathbf{P}_{\text{MDS}}$ | $\mathbf{F}_{\text{out}}^{\text{deconv3}}$ | $\mathbf{P}_{\text{out}}$ |
| Number of rows | $N_{\text{in}}/8$ | 256 | 512 | $1024 + N_{\text{in}}/4$ | 2048 | 4096 | 16,384 |
| Number of columns | 256 | 256 | 256 | 3 | 3 | 3 | 3 |

Note that the input to the architecture is the point cloud $\mathbf{P}_{\text{in}}$ with a dimension of $N_{\text{in}} \times 3$ while the output is another point cloud $\mathbf{P}_{\text{out}}$ with $16{,}384 \times 3$

**Skip Connection with Feature Transform** We bridge the encoder and decoder with a skip connection to build a U-Net-like (Çiçek et al. 2016; Yang et al. 2018a) structure. This connection links the results of conv1, denoted as $\mathbf{F}_{\text{out}}^{\text{conv1}}$, to the results of deconv1, denoted as $\mathbf{F}_{\text{out}}^{\text{deconv1}}$. However, instead of simply concatenating them, we introduce a square matrix $\mathbf{R}$ that transforms the features from the encoder as $\mathbf{F}_{\text{out}}^{\text{conv1}}\mathbf{R}$. Note that the multiplication by the transform is on the right side because the points are arranged row-wise in $\mathbf{P}_{\text{in}}$, which implies that the feature vectors are also arranged row-wise. Subsequently, we concatenate the two matrices into $[\mathbf{F}_{\text{out}}^{\text{deconv1}}, \mathbf{F}_{\text{out}}^{\text{conv1}}\mathbf{R}]$ that serves as the input to the feature extraction module, producing $\mathbf{F}_{\text{out}}^{\text{deconv2}}$.

In order to avoid randomly large values in the transformations and attain numerical stability during training, we regularize the transformation matrix to be orthonormal such that all elements are between $[-1, 1]$ and it mathematically satisfies $\mathbf{R}\mathbf{R}^\top = \mathbf{I}$ where $\mathbf{I}$ is an identity matrix. Geometrically, the regularizer imposes to rotate the features by $\mathbf{R}$.

**Minimum Density Sampling** Since the number of points in the input cloud vary, the results of deconv2 would also produce a varying number of points, i.e. with $1024 + \frac{N_{\text{in}}}{4}$ points from Table 1, since it depends on the input dimension. Thus, we include a Minimum Density Sampling (MDS) (Liu et al. 2020) in the decoder to standardize the output to a coarse resolution of 2048 points. The coarse resolution is then refined with two deconvolutional operations to 16,384 points. The motivation of adding the MDS is to help the final deconvolutional layers to converge faster during training. Later in Sect. 6, we investigate further the differences between the point clouds from the skip-connection as well as the coarse and fine as illustrated in Fig. 5.

## 5 Loss Functions

Since the main goal here is point cloud completion (Groueix et al. 2018; Yang et al. 2018b; Yuan et al. 2018), we first analyse whether the predicted point feature $\mathbf{P}_{\text{out}}$ matches the given ground truth $\mathbf{P}_{\text{gt}}$ through the Chamfer distance

$$\mathcal{L}_{\text{complete}} = \text{Chamfer}(\mathbf{P}_{\text{out}}, \mathbf{P}_{\text{gt}}) \,. \tag{2}$$

Furthermore, we optimize our architecture with two sets of loss functions that are related to the feature extraction module for all the convolution and deconvolution layers in the architecture from Sect. 3.4 as well as the skip connection with the feature transform from Sect. 4.

### 5.1 Optimizing the Feature Extraction Module

For the feature extraction module that utilizes SoftPool features, we adopt the same loss terms as in Wang et al. (2020b), where their main objective is to optimize the distribution of the features across different regions.

**Intra-regional Entropy** The ideal case for the feature vector $\mathbf{f}_i$ is a one-hot code, i.e. each vector gets assigned to only one region. To accomplish this goal, we measure the probability of $\mathbf{f}_i$ belonging to region $k$ in all $N_s$ regions by directly applying the softmax on the elements of the vector as

$$P(\mathbf{f}_i, k) = \frac{e^{\mathbf{f}_i[k]}}{\sum_{j=1}^{N_s} e^{\mathbf{f}_i[j]}} \,. \tag{3}$$

This implies that P is maximized when $\mathbf{f}_i$ is a one-hot code, with the $k$-th element equal to one. However, in presence of multiple peaks in the vector, $P(\mathbf{f}_i, k)$ might decrease significantly. Therefore, by taking the entropy into account, the

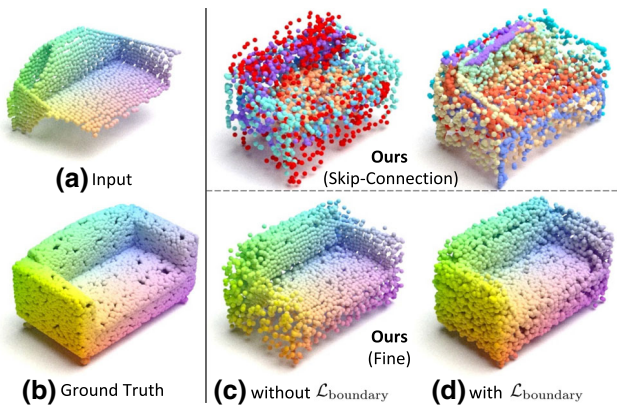

**Fig. 6** Our object completion results with and without the influence of $\mathcal{L}_{\text{boundary}}$

intra-regional loss function

$$\mathcal{L}_{\text{intra}} = -\frac{1}{N_{\text{in}}}\frac{1}{B}\sum_{i=1}^{N_{\text{in}}}\sum_{j=1}^{B}\sum_{k=1}^{N_s} P(\mathbf{f}_i, k) \log P(\mathbf{f}_i, k) \quad (4)$$

where $B$ is the batch size, tries to enforce the feature vector to have one peak so that it confidently falls into just one region.

**Inter-regional Entropy** The drawback of $\mathcal{L}_{\text{intra}}$ is that all feature vectors have the same peak at the $k$-th element. Looking at a more holistic perspective, the inter-regional loss function aims at distributing the features across different regions. It relies on maximizing the regional entropy

$$\mathcal{E}_r = -\frac{1}{B}\sum_{j=1}^{B}\sum_{k=1}^{N_s} \bar{P}_k \cdot \log \bar{P}_k \quad (5)$$

given that

$$\bar{P}_k = \frac{1}{N_{\text{in}}}\sum_{i=1}^{N_{\text{in}}} P(\mathbf{f}_i, k) \,. \quad (6)$$

We can then define the loss function as

$$\mathcal{L}_{\text{inter}} = \log(N_s) - \mathcal{E}_r \quad (7)$$

since the upper-bound of $\mathcal{E}_r$ is computed as $-\log \frac{1}{N_s}$ or simply $\log(N_s)$.

**Boundary Overlap Minimization** In addition to optimizing the holistic distribution of the points, we also incorporate a loss function that is applied on pairs of regions $i$ and $j$. We collect a set of points $\mathcal{B}_j^i$ from region $i$ with activations of region $j$ larger than a threshold $\tau$, i.e. set to 0.3. Similarly, we also take the inverse $\mathcal{B}_i^j$. Consequently, we squeeze the overlaps between the two regions.

By minimizing the Chamfer distance between $\mathcal{B}_j^i$ and $\mathcal{B}_i^j$, we obtain the loss

$$\mathcal{L}_{\text{boundary}} = \sum_{i=1}^{N_s}\sum_{j=i}^{N_s} \text{Chamfer}(\mathcal{B}_i^j, \mathcal{B}_j^i) \quad (8)$$

that tries to make the overlapping sets of points smaller, ideally down to just a line. In Fig. 6, we visualize the difference of optimizing with and without $\mathcal{L}_{\text{boundary}}$, where the distribution of the point cloud is less noisy on the occluded regions such as the armrest.

Notably, this loss function is general enough to be effectively applied also on other methods that rely on a subdivision of the point cloud into different regions, such as AtlasNet (Groueix et al. 2018), PCN (Yuan et al. 2018) and MSN (Liu et al. 2020). In Sect. 7.2, we formally evaluate these methods with and without $\mathcal{L}_{\text{boundary}}$.

**Feature Duplicate Minimization** The last loss term

$$\mathcal{L}_{\text{preserve}} = \text{Earth-moving}(\hat{\mathbf{F}}, \mathbf{F}) \quad (9)$$

imposes that the resulting truncated SoftPool feature $\hat{\mathbf{F}}$ takes most of the features from original $\mathbf{F}$ so that it avoids duplicates. To make the earth moving distance (Li et al. 2013) more efficient, 256 vectors are randomly selected from $\mathbf{F}$ and $\hat{\mathbf{F}}$. In practice, Fig. 7 visualizes the effects of $\mathcal{L}_{\text{preserve}}$ in the reconstruction, where lower weights of this loss produce a large hole, while incorporating this loss builds a point cloud with similar densities.

### 5.2 Optimizing the Skip Connection

We first visualize a subset of the architecture and focus on the skip connection as shown in Fig. 8. Here, we define $\mathbf{P}_{\text{partial}}$ as the partial reconstruction on $\mathbf{F}_{\text{out}}^{\text{deconv2}}$ contributed by the skip connection with the feature transform. However, note that $\mathbf{P}_{\text{partial}}$ is not a subset of $\mathbf{F}_{\text{out}}^{\text{deconv2}}$. It is produced by taking the input point cloud through conv1, feature transform and deconv2.

Since the skip connection aims to maintain the given input structure, we define a loss function that acts as an auto-encoder such that

$$\mathcal{L}_{\text{skip}} = \text{Chamfer}(\mathbf{P}_{\text{partial}}, \mathbf{P}_{\text{in}}) \,. \quad (10)$$

In addition, based on Sect. 4, we regularize the values in the feature transform such that

$$\mathcal{L}_{\mathbf{R}} = \|\mathbf{R}\mathbf{R}^\top - \mathbf{I}\|^2 \quad (11)$$

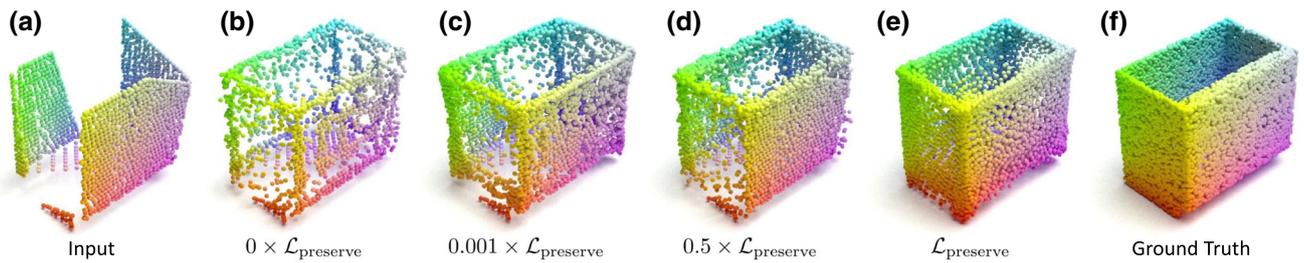

**Fig. 7** Our object completion results while increasing the weight of $\mathcal{L}_{\text{preserve}}$

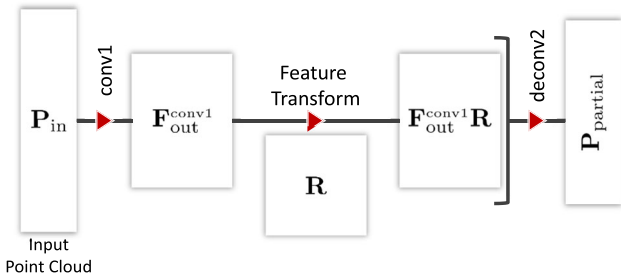

**Fig. 8** Subset of the architecture that focuses on the skip connection

makes **R** orthonormal.

### 5.3 Discriminative Training

Recognizing the advantages from TreeGAN (Shu et al. 2019), we also investigate applying discriminative training conditions on the input partial scan $\mathbf{P}_{\text{in}}$. In this case, we first introduce the conditional feature maps $\mathbf{P}_{\text{out}}|\mathbf{P}_{\text{in}}$ and $\mathbf{P}_{\text{gt}}|\mathbf{P}_{\text{in}}$ by concatenating them along the point axis. We build our discriminator $\mathcal{D}$ with the same parametric model proposed in Shu et al. (2019). By restricting the output of the discriminator to a range between 0 and 1, we can then apply

$$\mathcal{L}_{\text{infer}} = -\log(\mathcal{D}(\mathbf{P}_{\text{out}}|\mathbf{P}_{\text{in}})) ,  \qquad (12)$$

to optimize our completion architecture while

$$\mathcal{L}_{\text{discri}} = -\log(\mathcal{D}(\mathbf{P}_{\text{gt}}|\mathbf{P}_{\text{in}})) - \log(1 - \mathcal{D}(\mathbf{P}_{\text{out}}|\mathbf{P}_{\text{in}})) \qquad (13)$$

to optimize the discriminator $\mathcal{D}$. In practice, we impose the loss functions in (12) and (13) alternatively in order to optimize the completion architecture and the discriminators separately.

## 6 Experiments

For all evaluations, we train our model with an NVIDIA Titan V and parameterize it with a batch size of 8. Moreover, we apply the Leaky ReLU with a negative slope of 0.2 on the output of each regional convolution.

### 6.1 Completion on ShapeNet

We evaluate the performance of the geometric completion of a single object on the ShapeNet (Chang et al. 2015) database where they have the point clouds of the partial scanning as input and the corresponding ground truth completed shape. To make it comparable to other approaches, we adopt the standard 8 category evaluation (Yuan et al. 2018) for a single object completion. Both sampled from ShapeNet meshes, PCN (Yuan et al. 2018) and TopNet (Tchapmi et al. 2019) supplement two set of datasets individually for low and high resolutions evaluation, which contain 2048 and 16,384 points, respectively, where the inputs are provided with 2048 points. Notice that the low resolution dataset provided by TopNet is also commonly referred to Completion3D benchmark. Since previous works report their results in terms of L1/L2 metric of the Chamfer distance separately, we also report our results in both resolutions (2048 and 16,384) and metrics (L1 and L2).

We compare against state-of-the-art point cloud completion approaches such as PCN (Yuan et al. 2018), FoldingNet (Yang et al. 2018b), AtlasNet (Groueix et al. 2018), PointNet++ (Qi et al. 2017b), MSN (Liu et al. 2020) and GRNet (Xie et al. 2020b). To show the advantages over volumetric completion, we also compare against 3D-EPN (Dai et al. 2017) and ForkNet (Wang et al. 2019b) with an output resolution of $64 \times 64 \times 64$. As for point cloud resolutions, PCN (Yuan et al. 2018), GRNet (Xie et al. 2020b) and SoftPoolNet (Wang et al. 2020b) report the best performance with 16,384 points while MSN (Liu et al. 2020) presents their final output resolution with 8192 points. Aiming at a fair numerical comparison at different resolutions, we modify the last layers of these architectures so as to attain the same resolution for all methods.

**Low Resolution** At low resolution, we achieve competitive results, attaining the $0.13 \times 10^{-4}$ from PMP-Net (Wen et al. 2020b) with the L2-Chamfer distance in Table 2, while we achieve state-of-the-art results when evaluating on the L1-Chamfer distance in Table 3.

**Table 2** Evaluation on the object completion based on the Chamfer distance trained with L2 distance (multiplied by $10^4$) with the output resolution of 2048

| Method | Plane | Cabinet | Car | Chair | Lamp | Sofa | Table | Vessel | Avg. |
|---|---|---|---|---|---|---|---|---|---|
| Completion3D (Tchapmi et al. 2019) benchmark, Output Resolution = 2048, L2 metric | | | | | | | | | |
| FoldingNet (Yang et al. 2018b) | 12.83 | 23.01 | 14.88 | 25.69 | 21.79 | 21.31 | 20.71 | 11.51 | 19.07 |
| PointSetVoting (Zhang et al. 2020a) | 6.88 | 21.18 | 15.78 | 22.54 | 18.78 | 28.39 | 19.96 | 11.16 | 18.18 |
| AtlasNet (Groueix et al. 2018) | 10.36 | 23.40 | 13.40 | 24.16 | 20.24 | 20.82 | 17.52 | 11.62 | 17.77 |
| PCN (Yuan et al. 2018) | 9.79 | 22.70 | 12.43 | 25.14 | 22.72 | 20.26 | 20.27 | 11.73 | 18.22 |
| TopNet (Tchapmi et al. 2019) | 7.32 | 18.77 | 12.88 | 19.82 | 14.60 | 16.29 | 14.89 | 8.82 | 14.25 |
| GRNet (Xie et al. 2020b) | 6.13 | 16.90 | 8.27 | 12.23 | 10.22 | 14.93 | 10.08 | 5.86 | 10.64 |
| SA-Net (Wen et al. 2020a) | 5.27 | 14.45 | 7.78 | 13.67 | 13.53 | 14.22 | 11.75 | 8.84 | 11.22 |
| PMP-Net (Wen et al. 2020b) | **3.99** | **14.70** | 8.55 | **10.21** | **9.27** | **12.43** | **8.51** | **5.77** | **9.23** |
| SoftPoolNet (Wang et al. 2020b) | 6.39 | 17.26 | 8.72 | 13.16 | 10.78 | 14.95 | 11.01 | 6.26 | 11.07 |
| Ours | 4.59 | 15.82 | 6.78 | 11.41 | 8.82 | 13.37 | 9.15 | 4.93 | 9.36 |
| Without skip-connection | 4.63 | 16.35 | 9.10 | 13.40 | 10.55 | 13.85 | 10.90 | 6.23 | 10.63 |
| Without $\mathcal{D}$ | 5.07 | 16.12 | 6.86 | 11.56 | 8.88 | 13.67 | 9.21 | 5.33 | 9.59 |
| Without $\mathcal{L}_R$ | 5.38 | 17.04 | 9.93 | 14.13 | 11.35 | 14.52 | 11.63 | 6.81 | 11.35 |

Bold indicates the best performance achieved in certain column

**Table 3** Evaluation on the object completion based on the Chamfer distance trained with L1 distance (multiplied by $10^4$) with the output resolution of 2048

| Method | Plane | Cabinet | Car | Chair | Lamp | Sofa | Table | Vessel | Avg. |
|---|---|---|---|---|---|---|---|---|---|
| Completion3D (Tchapmi et al. 2019) benchmark, output resolution = 2048, L1 metric | | | | | | | | | |
| FoldingNet (Yang et al. 2018b) | 11.18 | 20.15 | 13.25 | 21.48 | 18.19 | 19.09 | 17.80 | 10.69 | 16.48 |
| AtlasNet (Groueix et al. 2018) | 10.37 | 23.40 | 13.41 | 24.16 | 20.24 | 20.82 | 17.52 | 11.62 | 17.69 |
| AtlasNet + $\mathcal{L}_{boundary}$ | 9.25 | 22.51 | 12.12 | 22.64 | 18.82 | 19.11 | 16.50 | 11.53 | 16.56 |
| PCN (Yuan et al. 2018) | 8.09 | 18.32 | 10.53 | 19.33 | 18.52 | 16.44 | 16.34 | 10.21 | 14.72 |
| PCN + $\mathcal{L}_{boundary}$ | 6.39 | 16.32 | 9.30 | 18.61 | 16.72 | 16.28 | 15.29 | 9.00 | 13.49 |
| TopNet (Tchapmi et al. 2019) | 5.50 | 12.02 | 8.90 | 12.56 | 9.54 | 12.20 | 9.57 | 7.51 | 9.72 |
| SA-Net (Wen et al. 2020a) | **2.18** | **9.11** | **5.56** | **8.94** | 9.98 | **7.83** | 9.94 | 7.23 | 7.74 |
| SoftPoolNet (Wang et al. 2020b) | 4.76 | 10.29 | 7.63 | 11.23 | 8.97 | 10.08 | 7.13 | 6.38 | 8.31 |
| Ours | 3.50 | 9.95 | 7.01 | 10.48 | **8.45** | 8.86 | **5.99** | **5.60** | **7.48** |
| Without skip-connection | 4.29 | 10.24 | 7.76 | 11.10 | 9.13 | 9.72 | 6.33 | 6.46 | 8.13 |
| Without $\mathcal{D}$ | 3.72 | 10.07 | 7.23 | 10.76 | 8.50 | 9.15 | 6.10 | 5.92 | 7.68 |
| Without $\mathcal{L}_R$ | 4.68 | 10.54 | 8.06 | 11.42 | 9.45 | 10.03 | 6.70 | 6.77 | 8.46 |
| Without $\mathcal{L}_{inter}$ | 9.81 | 19.49 | 13.24 | 18.20 | 16.83 | 17.00 | 15.64 | 7.16 | 14.67 |
| Without $\mathcal{L}_{intra}$ | 3.70 | 15.93 | 10.78 | 12.97 | 12.89 | 11.79 | 11.33 | 5.60 | 10.62 |
| Without $\mathcal{L}_{inter}, \mathcal{L}_{intra}$ | 9.07 | 19.62 | 14.31 | 19.17 | 16.46 | 17.82 | 14.34 | 7.60 | 14.80 |
| Without $\mathcal{L}_{boundary}$ | 4.71 | 10.43 | 8.14 | 11.27 | 9.27 | 10.57 | 7.43 | 7.36 | 8.65 |
| Without $\mathcal{L}_{preserve}$ | 7.43 | 18.84 | 11.58 | 15.68 | 17.38 | 17.53 | 14.46 | 6.49 | 13.68 |

Bold indicates the best performance achieved in certain column

**High Resolution** We achieve the best results on most objects with the high resolution as presented in Tables 4 and 5 with $8.31 \times 10^{-3}$ and $2.55 \times 10^{-3}$, respectively. Table 5 also shows that volumetric approaches like 3D-EPN (Dai et al. 2017) and ForkNet (Wang et al. 2019b) having large issues when evaluated in Chamfer distance because the converted point clouds from the fixed volumetric grids are at much smaller local resolutions.

**Validating with F-Score@1%** Since the Chamfer distance hardly reflect the errors in the local geometry as suggested in Tatarchenko et al. (2019), the evaluation in GRNet (Xie et al. 2020b) uses the metric F-Score@1% that computes the F-Score after matching the predicted point cloud to the ground truth with a distance threshold of 1% of the side length of the reconstructed volume. The evaluations on reconstructing higher resolutions are reported in Tables 6 and 7 on ShapeNet

**Table 4** Evaluation on the object completion based on the Chamfer distance trained with L1 distance (multiplied by $10^3$) with the output resolution of 16,384

| Method | Plane | Cabinet | Car | Chair | Lamp | Sofa | Table | Vessel | Avg. |
|---|---|---|---|---|---|---|---|---|---|
| PCN (Yuan et al. 2018) dataset, Output Resolution = 16,384, L1 metric | | | | | | | | | |
| 3D-EPN (Dai et al. 2017) | 13.16 | 21.80 | 20.31 | 18.81 | 25.75 | 21.09 | 21.72 | 18.54 | 20.15 |
| ForkNet (Wang et al. 2019b) | 9.08 | 14.22 | 11.65 | 12.18 | 17.24 | 14.22 | 11.51 | 12.66 | 12.85 |
| PointNet++ (Qi et al. 2017b) | 10.30 | 14.74 | 12.19 | 15.78 | 17.62 | 16.18 | 11.68 | 13.52 | 14.00 |
| FoldingNet (Yang et al. 2018b) | 9.49 | 15.80 | 12.61 | 15.55 | 16.41 | 15.97 | 13.65 | 14.99 | 14.31 |
| AtlasNet (Groueix et al. 2018) | 6.37 | 11.94 | 10.11 | 12.06 | 12.37 | 12.99 | 10.33 | 10.61 | 10.85 |
| TopNet (Tchapmi et al. 2019) | 7.61 | 13.31 | 10.90 | 13.82 | 14.44 | 14.78 | 11.22 | 11.12 | 12.15 |
| PCN (Yuan et al. 2018) | 5.50 | 10.63 | 8.70 | 11.00 | 11.34 | 11.68 | 8.59 | 9.67 | 9.64 |
| PCN + $\mathcal{L}_{boundary}$ | 5.13 | 9.12 | 7.58 | 9.35 | 9.40 | 9.31 | 7.30 | 8.91 | 8.26 |
| MSN (Liu et al. 2020) | 5.60 | 11.96 | 10.78 | 10.62 | 10.71 | 11.90 | 8.70 | 9.49 | 9.97 |
| GRNet (Xie et al. 2020b) | 6.45 | 10.37 | 9.45 | 9.41 | 7.96 | 10.51 | 8.44 | 8.04 | 8.83 |
| PMP-Net (Wen et al. 2020b) | 5.65 | 11.24 | 9.64 | 9.51 | 6.95 | 10.83 | 8.72 | 7.25 | 8.73 |
| CRN (Wang et al. 2020a) | **4.79** | **9.97** | **8.31** | 9.49 | 8.94 | 10.69 | **7.81** | 8.05 | 8.51 |
| SoftPoolNet (Wang et al. 2020b) | 6.93 | 10.91 | 9.78 | 9.56 | 8.59 | 11.22 | 8.51 | 8.14 | 9.20 |
| Ours | 5.50 | 10.02 | 8.73 | **9.05** | 7.53 | 10.24 | 8.01 | **7.43** | **8.31** |
| Without skip-connection | 6.72 | 10.46 | 9.70 | 9.12 | 8.42 | 10.85 | 8.48 | 7.80 | 8.95 |
| Without $\mathcal{D}$ | 5.73 | 10.19 | 8.79 | 9.10 | 7.55 | 10.47 | 8.12 | 7.75 | 8.46 |
| Without $\mathcal{L}_R$ | 5.77 | 11.92 | 11.60 | 11.47 | 9.02 | 12.14 | 11.82 | 9.87 | 10.45 |

Bold indicates the best performance achieved in certain column

**Table 5** Evaluation on the object completion based on the Chamfer distance trained with L2 distance (multiplied by $10^3$) with the output resolution of 16,384

| Method | Plane | Cabinet | Car | Chair | Lamp | Sofa | Table | Vessel | Avg. |
|---|---|---|---|---|---|---|---|---|---|
| PCN (Yuan et al. 2018) dataset, Output Resolution = 16,384, L2 metric | | | | | | | | | |
| FoldingNet (Yang et al. 2018b) | 3.15 | 7.94 | 4.68 | 9.23 | 9.23 | 8.90 | 6.69 | 7.33 | 7.14 |
| FoldingNet + *SoftPool++* | 3.02 | 7.86 | 4.50 | 9.07 | 9.03 | 8.69 | 6.49 | 7.31 | 7.00 |
| AtlasNet (Groueix et al. 2018) | 1.75 | 5.10 | 3.24 | 5.23 | 6.34 | 5.99 | 4.36 | 4.18 | 4.52 |
| TopNet (Tchapmi et al. 2019) | 2.15 | 5.62 | 3.51 | 6.35 | 7.50 | 6.95 | 4.78 | 4.36 | 5.15 |
| NSFA (Zhang et al. 2020b) | 1.75 | 5.31 | 3.43 | 5.01 | 4.73 | 6.41 | 4.00 | 3.56 | 4.28 |
| PCN (Yuan et al. 2018) | 1.40 | 4.45 | 2.45 | 4.84 | 6.24 | 5.13 | 3.57 | 4.06 | 4.02 |
| PCN + *SoftPool++* | **1.10** | 4.37 | **2.40** | 4.81 | 5.67 | 4.70 | 3.41 | 3.82 | 3.79 |
| MSN (Liu et al. 2020) | 1.54 | 7.25 | 4.71 | 4.54 | 6.48 | 5.89 | 3.80 | 3.85 | 4.76 |
| MSN + *SoftPool++* | 1.13 | 7.24 | 4.64 | 4.21 | 6.28 | 5.83 | 3.57 | 3.45 | 4.54 |
| PF-Net (Huang et al. 2020) | 1.55 | 4.43 | 3.12 | 3.96 | 4.21 | 5.87 | 3.35 | 3.89 | 3.80 |
| CRN (Wang et al. 2020a) | 1.46 | 4.21 | 2.97 | 3.24 | 5.16 | 5.01 | 3.99 | 3.96 | 3.75 |
| GRNet (Xie et al. 2020b) | 1.53 | 3.62 | 2.75 | **2.95** | **2.65** | 3.61 | 2.55 | 2.12 | 2.72 |
| SoftPoolNet (Wang et al. 2020b) | 1.63 | 3.79 | 3.05 | 3.27 | 2.95 | 3.78 | 2.59 | 2.25 | 2.91 |
| Ours | 1.27 | **3.43** | 2.65 | 2.98 | 2.67 | **3.38** | **2.27** | **1.85** | **2.55** |
| Without skip-connection | 1.53 | 3.75 | 2.96 | 3.15 | 2.90 | 3.59 | 2.35 | 1.96 | 2.77 |
| Without $\mathcal{D}$ | 1.37 | 3.59 | 2.78 | 3.13 | 2.74 | 3.51 | 2.43 | 2.03 | 2.69 |
| Without $\mathcal{L}_R$ | 1.42 | 4.74 | 2.91 | 4.63 | 3.66 | 4.14 | 2.83 | 2.29 | 3.33 |

Bold indicates the best performance achieved in certain column

**Table 6** Evaluation on the object completion based on the F-Score@1% trained with L2 Chamfer distance and the output resolution of 16,384

| Method | Plane | Cabinet | Car | Chair | Lamp | Sofa | Table | Vessel | Avg. |
|---|---|---|---|---|---|---|---|---|---|
| PCN (Yuan et al. 2018) dataset, output resolution = 16,384, F-Score@1% | | | | | | | | | |
| FoldingNet (Yang et al. 2018b) | 0.642 | 0.237 | 0.382 | 0.236 | 0.219 | 0.197 | 0.361 | 0.299 | 0.322 |
| FoldingNet + *SoftPool++* | 0.687 | 0.347 | 0.455 | 0.237 | 0.236 | 0.257 | 0.377 | 0.428 | 0.378 |
| AtlasNet (Groueix et al. 2018) | 0.845 | 0.552 | 0.630 | 0.552 | 0.565 | 0.500 | 0.660 | 0.624 | 0.616 |
| TopNet (Tchapmi et al. 2019) | 0.771 | 0.404 | 0.544 | 0.413 | 0.408 | 0.350 | 0.572 | 0.560 | 0.503 |
| PCN (Yuan et al. 2018) | 0.881 | 0.651 | 0.725 | 0.625 | 0.638 | 0.581 | 0.765 | 0.697 | 0.695 |
| PCN + *SoftPool++* | 0.880 | 0.671 | **0.777** | 0.723 | 0.755 | 0.578 | 0.819 | 0.661 | 0.733 |
| MSN (Liu et al. 2020) | 0.885 | 0.644 | 0.665 | 0.657 | 0.699 | 0.604 | 0.782 | 0.708 | 0.705 |
| MSN + *SoftPool++* | **0.903** | 0.727 | 0.721 | **0.736** | 0.718 | 0.633 | 0.796 | 0.750 | 0.748 |
| GRNet (Xie et al. 2020b) | 0.843 | 0.618 | 0.682 | 0.673 | 0.761 | 0.605 | 0.751 | 0.750 | 0.708 |
| SoftPoolNet (Wang et al. 2020b) | 0.831 | 0.605 | 0.685 | 0.649 | 0.715 | 0.601 | 0.746 | 0.721 | 0.694 |
| Ours | 0.867 | **0.693** | 0.706 | 0.712 | **0.794** | **0.689** | **0.825** | **0.804** | **0.761** |
| Without skip-connection | 0.836 | 0.658 | 0.670 | 0.671 | 0.753 | 0.652 | 0.753 | 0.791 | 0.723 |
| Without $\mathcal{D}$ | 0.843 | 0.672 | 0.700 | 0.686 | 0.767 | 0.653 | 0.768 | 0.796 | 0.736 |
| Without $\mathcal{L}_R$ | 0.824 | 0.634 | 0.593 | 0.670 | 0.695 | 0.575 | 0.686 | 0.755 | 0.679 |

Bold indicates the best performance achieved in certain column objects provided by the Completion3D (Tchapmi et al. 2019) and MVP (Pan et al. 2021), respectively. Here, the average F-Score with SoftPool++ outperforms the other methods. The tables also validate the benefit of our individual contributions in the overall result. In addition, Table 7 shows that, by applying our SoftPool++ module on the variational coarse sub-architecture of VRCNet (Pan et al. 2021), the average performance of the fine reconstruction reached the state-of-the-art with the improvement from 78.1 to 79.9%.

*Advantage over SoftPoolNet* Wang et al. (2020b).

Compared to SoftPoolNet (Wang et al. 2020b), our contributions in the proposed SoftPool++ features improve (Wang et al. 2020b) by $0.83 \times 10^{-4}$ in the L1 Chamfer distance and $1.71 \times 10^{-4}$ for L2. Strikingly, even without the skip connections, we have already outperformed SoftPoolNet (Wang et al. 2020b). This then demonstrate the strength of the proposed SoftPool++ over (Wang et al. 2020b).

Moreover, the results from high resolution reconstruction also validates our conclusion when evaluating against SoftPoolNet (Wang et al. 2020b). With or without the skip connections, our SoftPool++ performs better than (Wang et al. 2020b).

### 6.2 Qualitative Evaluation

Similar to Sect. 6.1, the objects in this section are also trained from and evaluated on ShapeNet (Chang et al. 2015). However, for the qualitative results in Fig. 9, we show the results in the original points resolution specified in their respective methods.

*Comparison against PointNet* (Qi et al. 2017a) *feature.*

From Fig. 9, the max-pooling operation from the PointNet (Qi et al. 2017a) feature is embedded in FoldingNet (Yang et al. 2018b), PCN (Yuan et al. 2018) and MSN (Liu et al. 2020). We noticed that these methods are either over-smoothens the reconstruction or start introducing noise.

On one hand, FoldingNet (Yang et al. 2018b) and PCN (Yuan et al. 2018) smoothens out the reconstruction so that the fine details such as the armrest of the chair are no longer visible and the wheels of the car are no longer separated. On the other, MSN (Liu et al. 2020) tries to reconstruct the finer details but produces a noisy point cloud. Contrary to these methods, we achieve a smoother surface reconstruction with with visible geometric details of the object like the armrest and the wheels.

**Advantage of Skip Connections** We also explore the combination of 3D-GCN (Lin et al. 2020) and TreeGAN (Shu et al. 2019) that uses graph convolutions in an encoder–decoder architecture. Its latent feature is presented as a vector with a length of 1024. Without the skip connection, several inconsistencies emerge. For instance, the shape of the boat is slimmer than the ground truth while one dimension of the bookshelf is thicker. These information are part of the input but are not propagated to the output.

Among these methods, GRNet (Xie et al. 2020b) achieves similar quantitative results compared to our approach in Table 5. They also build skip connections between their encoder and decoder. However, as input to the architecture, they first voxelize the input point cloud. After going through the encoder–decoder, they convert the 3D grid back to point cloud. Due to the discretization of the point cloud, this affects the results of GRNet (Xie et al. 2020b). It fails to reconstruct

**Table 7** Evaluation on the object completion based on the F-Score@1% trained with L2 Chamfer distance and the output resolution of 16,384

| Method | Plane | Cabinet | Car | Chair | Lamp | Sofa | Table | Vessel | Avg. |
| --- | --- | --- | --- | --- | --- | --- | --- | --- | --- |
| MVP (Pan et al. 2021) dataset, Output Resolution = 16,384, F-Score@1% | | | | | | | | | |
| TopNet (Tchapmi et al. 2019) | 0.789 | 0.621 | 0.612 | 0.443 | 0.387 | 0.506 | 0.639 | 0.609 | 0.576 |
| PCN (Yuan et al. 2018) | 0.816 | 0.614 | 0.686 | 0.517 | 0.455 | 0.552 | 0.646 | 0.628 | 0.614 |
| PCN + *SoftPool++* | 0.853 | 0.643 | 0.729 | 0.563 | 0.472 | 0.566 | 0.670 | 0.643 | 0.642 |
| MSN (Liu et al. 2020) | 0.879 | 0.692 | 0.693 | 0.599 | 0.604 | 0.627 | 0.730 | 0.696 | 0.690 |
| MSN + *SoftPool++* | 0.914 | 0.717 | 0.727 | 0.620 | 0.638 | 0.649 | 0.765 | 0.726 | 0.719 |
| GRNet (Xie et al. 2020b) | 0.853 | 0.578 | 0.646 | 0.635 | 0.710 | 0.580 | 0.690 | 0.723 | 0.677 |
| ECG (Pan 2020) | 0.906 | 0.680 | 0.716 | 0.683 | 0.734 | 0.651 | 0.766 | 0.753 | 0.736 |
| NSFA (Zhang et al. 2020b) | 0.903 | 0.694 | 0.721 | 0.737 | 0.783 | 0.705 | 0.817 | 0.799 | 0.770 |
| CRN (Wang et al. 2020a) | 0.898 | 0.688 | 0.725 | 0.670 | 0.681 | 0.641 | 0.748 | 0.742 | 0.724 |
| VRCNet (Pan et al. 2021) | 0.928 | 0.721 | 0.756 | 0.743 | 0.789 | 0.696 | 0.813 | 0.800 | 0.781 |
| VRCNet + *SoftPool++* | **0.947** | **0.745** | **0.768** | **0.759** | **0.810** | **0.720** | **0.829** | **0.813** | **0.799** |
| PoinTr (Yu et al. 2021) | 0.888 | 0.681 | 0.716 | 0.703 | 0.749 | 0.656 | 0.773 | 0.760 | 0.741 |
| SoftPoolNet (Wang et al. 2020b) | 0.843 | 0.568 | 0.636 | 0.623 | 0.698 | 0.568 | 0.680 | 0.710 | 0.666 |
| Ours | 0.862 | 0.622 | 0.704 | 0.695 | 0.783 | 0.649 | 0.776 | 0.778 | 0.734 |
| Without skip-connection | 0.862 | 0.555 | 0.648 | 0.652 | 0.716 | 0.603 | 0.703 | 0.719 | 0.682 |
| Without $\mathcal{D}$ | 0.856 | 0.624 | 0.666 | 0.664 | 0.732 | 0.622 | 0.738 | 0.770 | 0.709 |
| Without $\mathcal{L}_R$ | 0.822 | 0.488 | 0.602 | 0.573 | 0.661 | 0.500 | 0.667 | 0.696 | 0.626 |

Bold indicates the best performance achieved in certain column thin structures like the antenna on the boat and the vertical stabilizers of the jet. In addition, it tried to fill up the hole in the box which should have remained empty. In contrast, our method that processes directly on the point cloud can handle these cases.

*Improvements from SoftPoolNet* (Wang et al. 2020b). Moreover, we compared the proposed method against the previous SoftPoolNet (Wang et al. 2020b) to reveal the advantages of our novel approach. From Fig. 9, while the previous method fails to reconstruct the four corners of the box and the wheels of the jet, the new method is more consistent to the ground truth. Overall, our novel approach reconstructs sharper geometries with less noise and less holes.

**Other Methods** There have been some trend to re-purpose method that were originally tailored for semantic segmentation such as PointCNN (Li et al. 2018) to train for object completion. Since they both use point clouds, the intuition is to use the local convolutions from Li et al. (2018) to upsample the point cloud from its partial scan to its completed structure. Unfortunately, these methods fails to reconstruct the objects because it is not the intended purpose of the architecture—in semantic segmentation, their input and output point cloud remains the same.

### 6.3 Classification on ModelNet and PartNet

In addition to shape completion, we also evaluate our approach in terms of classification on the ModelNet10 (Zhirong et al. 2015), ModelNet40 (Zhirong et al. 2015) and PartNet (Mo et al. 2019) datasets. Note that ModelNet40 contains 12,311 CAD models classified into 40 categories while PartNet contains 26,671 models with 24 categories.

Similar to the other approaches such as 3D-GAN (Wu et al. 2016), RS-DGCNN (Sauder et al. 2019), VConv-DAE (Sharma et al. 2016), FoldingNet (Yang et al. 2018b) and KCNet (Shen et al. 2018), we also implemented a self-supervised training to extract features from the input point cloud then a supervised training to train a linear Support Vector Machine (SVM) (Cortes and Vapnik 1995) to predict the categorical classification. The former relies on the 57,448 ShapeNet models (Chang et al. 2015) as its training dataset while the latter relies on ModelNet (Zhirong et al. 2015) and PartNet (Mo et al. 2019).

It is noteworthy to mention that there is a significant difference from RS-DGCNN (Sauder et al. 2019) in the details of the self-supervised training. On one hand, our method randomly subsamples the point cloud while, on the other, Sauder et al. (2019) includes an additional data augmentation step that randomly decomposes the 3D input structure into different parts then repositions these parts by translation. Since we did not include the additional augmentation from Sauder et al. (2019), our evaluation is a fair comparison against other methods.

The evaluation in Table 8 reports that our model outperforms the accuracy of RS-DGCNN (Sauder et al. 2019) by 4.11% on the ModelNet40 dataset, a sign of the higher descriptiveness in terms of categorical information. The

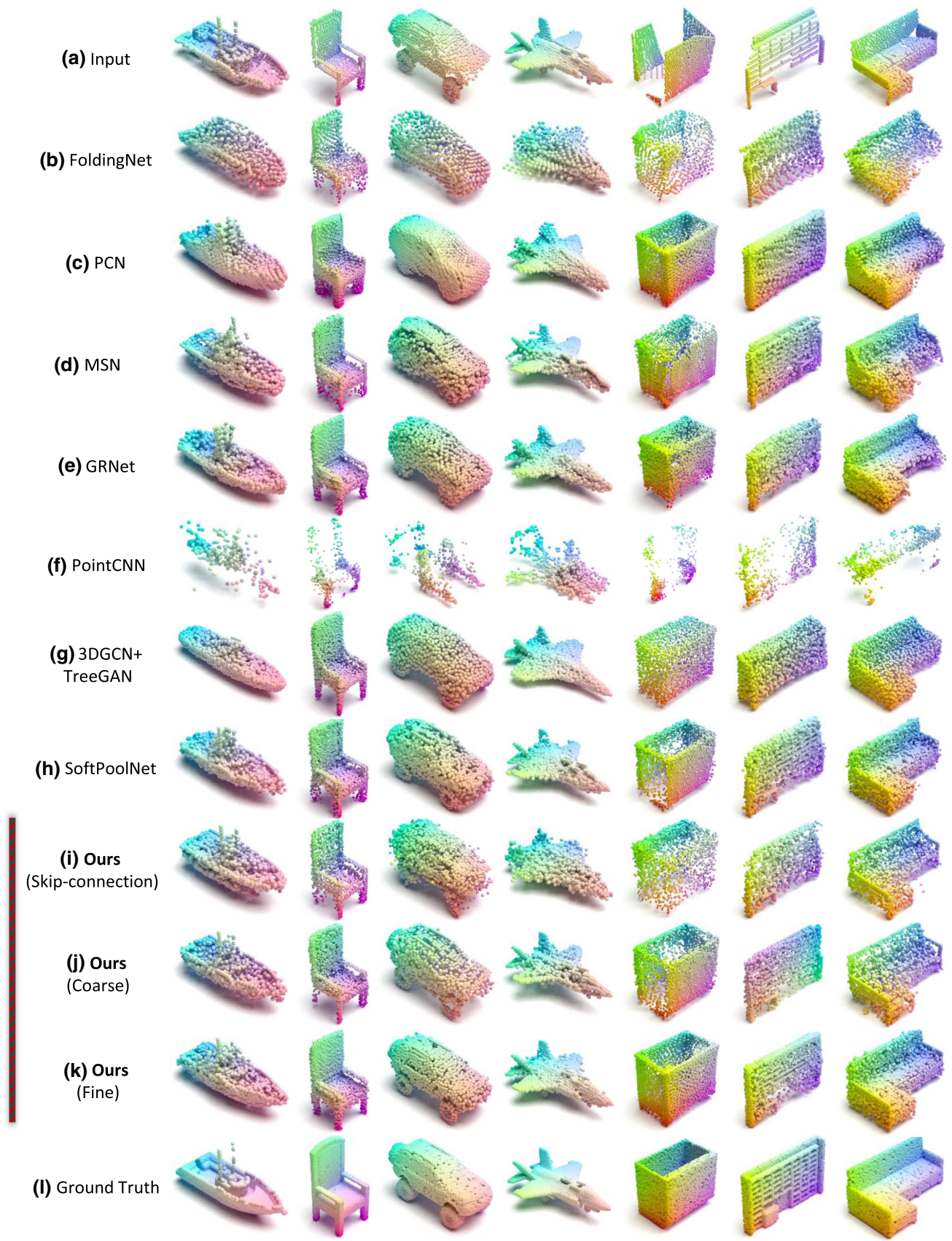

**Fig. 9** Qualitative results on the ShapeNet (Chang et al. 2015) dataset. Note that the three results from our method corresponds to different parts of the architecture as explained in Fig. 5. Here, (k) represents our final reconstruction

**Table 8** Object classification accuracy on ModelNet40 (Zhirong et al. 2015), ModelNet40 (Zhirong et al. 2015) and PartNet (Mo et al. 2019) datasets

| Method | ModelNet40 (%) | ModelNet10 (%) | PartNet (%) |
|---|---|---|---|
| VConv-DAE (Sharma et al. 2016) | 75.50 | 80.50 | – |
| 3D-GAN (Wu et al. 2016) | 83.30 | 91.00 | 74.23 |
| Latent-GAN | 85.70 | 95.30 | – |
| FoldingNet (Yang et al. 2018b) | 88.40 | 94.40 | – |
| VIP-GAN (Han et al. 2019) | 90.19 | 92.18 | – |
| RS-PointNet (Sauder et al. 2019) | 87.31 | 91.61 | 76.95 |
| RS-DGCNN (Sauder et al. 2019) | 90.64 | 94.52 | – |
| KCNet (Shen et al. 2018) | 91.0 | 94.4 | – |
| SoftPoolNet (Wang et al. 2020b) | 92.28 | 96.14 | 84.32 |
| Ours | **94.75** | **96.99** | **87.25** |
| Without skip-connection | 93.17 | 96.34 | 85.26 |
| Without $\mathcal{L}_{inter}$ | 91.23 | 95.11 | 83.07 |
| Without $\mathcal{L}_{intra}$ | 84.98 | 91.35 | 81.91 |
| Without $\mathcal{L}_{inter}$, $\mathcal{L}_{intra}$ | 84.21 | 91.77 | 80.55 |
| Without $\mathcal{L}_{boundary}$ | 89.70 | 94.14 | 82.39 |
| Without $\mathcal{L}_{preserve}$ | 87.84 | 93.15 | 81.32 |
| Without $\mathcal{L}_R$ | 79.22 | 85.40 | 76.48 |

Bold indicates the best performance achieved in certain column improvement of 2.47% from our approach compared to SoftPoolNet (Wang et al. 2020b) is also obvious, proving that the proposed SoftPool++ feature and skip-connection together are more advantageous for classification. Similar results are also obtained on ModelNet10 (Zhirong et al. 2015) and PartNet (Mo et al. 2019).

## 6.4 Efficiency

In addition to the evaluation in terms of shape completion and categorical classification, we also compare in Table 9 the properties of our model such as its memory footprint and inference speed, as well as the type of data being processed.

The cost of outperforming SoftPoolNet (Wang et al. 2020b) becomes evident on the memory footprint and the inference time. Compared to SoftPoolNet (Wang et al. 2020b), the memory footprint of our method is approximately doubled due the increase in the number of parameters from the multiple feature extraction modules in our architecture. This also triggers a larger inference time than SoftPoolNet (Wang et al. 2020b) from 0.04 to 0.11 seconds. A similar trend is associated to other approaches that divides the point cloud into regions such as AtlasNet (Groueix et al. 2018) and MSN (Liu et al. 2020), i.e. we achieve significantly higher accuracy in reconstruction but also increase the memory footprint and the inference time.

However, if we look at the overall data, we observe that the proposed method at 61.7MB consumes remarkably less memory than the other point cloud approaches such as GRNet (Xie et al. 2020b) at 293MB and PointCNN (Li et al. 2018) at 497MB, as well as the volumetric approaches such as 3D-EPN (Dai et al. 2017) at 420MB and ForkNet (Wang et al. 2019b) at 362MB. An important reason why their models are so large in memory usage is that 3D convolutions are applied in multiple layers of their architectures, while our approach is mainly composed of 2D convolutions only. Among those approaches with large memory consumption, GRNet (Xie et al. 2020b) is one of the top performers in point cloud completion. Since their architecture relies on volumetric grids where they convert the input point cloud to voxel grid then convert back to a point cloud, this affects not only their memory footprint but also their inference time, which is 8 times higher than ours.

Compared to approaches composed mainly of MLPs, our model reports a comparable size to PCN (Yuan et al. 2018) while having a faster inference time than MSN (Liu et al. 2020). The reason is that although our 2D convolution kernels introduces a additional dimensions, the newly added dimension $N_k$ of 32 is comparably much smaller than the feature dimension $N_f$ of 256 at which MLPs operates. Notably, approaches based on KNN search such as PointCNN (Li et al. 2018) and 3D-GCN (Lin et al. 2020) usually take much longer for inference.

## 7 Ablation Study

Based on the evaluation from ShapeNet (Chang et al. 2015), we further analyze our proposed method's behavior through an ablation study. In this section, we demonstrate the advantage of SoftPool++ over PontNet; expound on the claims of our loss function; investigate the value of the skip connection with feature transform in our architecture; and; delve deeper on what happens in the SoftPool++ module.

**Table 9** Overview of different object completion methods. Note that the inference time is represented by the amount of time to conduct inference on a single object

| Method | Size (MB) | Inference Time (s) | Core Operator | Data Type | With KNN |
|---|---|---|---|---|---|
| 3D-EPN (Dai et al. 2017) | 420.0 | – | 3D Conv | Voxels | No |
| ForkNet (Wang et al. 2019b) | 362.0 | – | 3D Conv | Voxels | No |
| GRNet (Xie et al. 2020b) | 293.0 | 0.88 | 3D Conv | Points | No |
| PointCNN (Li et al. 2018) | 497.0 | 1.20 | 3D Conv | Points | Yes |
| DeepSDF (Park et al. 2018) | 7.4 | 9.72 | MLP | SDF | No |
| FoldingNet (Yang et al. 2018b) | 19.2 | 0.05 | MLP | Points | No |
| AtlasNet (Groueix et al. 2018) | 2.0 | 0.32 | MLP | Points | No |
| PCN (Yuan et al. 2018) | 54.8 | 0.11 | MLP | Points | No |
| MSN (Liu et al. 2020) | 12.0 | 0.21 | MLP | Points | No |
| 3D-GCN (Lin et al. 2020) (coder) | 2.1 | 0.82 | 2D Conv | Points | Yes |
| SoftPoolNet (Wang et al. 2020b) | 37.2 | 0.04 | 2D Conv | Points | No |
| Ours | 61.7 | 0.11 | 2D Conv | Points | No |

### 7.1 Replacing PointNet with SoftPool++

In addition to the comparison in Table 5, we also evaluate the results by replacing the latent features in other point cloud completion approaches [i.e.FoldingNet (Yang et al. 2018b), MSN (Liu et al. 2020), and PCN (Yuan et al. 2018)] with our SoftPool++ features, while keeping their decoders unchanged. In this way, we have a one-to-one comparison of PointNet and SoftPool++ features.

Since these works depend on a PointNet features (having a dimensionality of 1024), we also build up our SoftPool++ features with the same size. Remarkably, the use of SoftPool++ features improves performance in all tested methods, i.e.the performance of FoldingNet (Yang et al. 2018b), PCN (Yuan et al. 2018) and MSN (Liu et al. 2020) improves respectively by $0.14 \times 10^{-3}$, $0.23 \times 10^{-3}$ and $0.22 \times 10^{-3}$.

### 7.2 Loss Functions

Tables 3 and 8 include an ablation study that investigates the effects of the individual loss functions from Sect. 5. For both experiments, we notice that all loss functions are critical to achieve state-of-the-art results. Note that we have shown in Fig. 6 and cabinet completion in Fig. 7 to demonstrate the contributions of $\mathcal{L}_{boundary}$ and $\mathcal{L}_{preserve}$ in the reconstruction. $\mathcal{L}_{boundary}$ *in other methods*. An interesting idea is the capacity of $\mathcal{L}_{boundary}$ to be integrated in other existing methods that join multiple deformed 2D patches together to form the final output. Since the patches in AtlasNet (Groueix et al. 2018), PCN (Yuan et al. 2018) and MSN (Liu et al. 2020) are frequently overlapping nearby patches, we tried to integrate $\mathcal{L}_{boundary}$ into their loss functions. Tables 4 and 3 evaluate this idea and prove that this activation helps FoldingNet (Yang et al. 2018b), PCN (Yuan et al. 2018) and AtlasNet (Groueix et al. 2018) perform better, improving the Chamfer distance with at least $1 \times 10^{-4}$ on the resolution of 2048 and $1 \times 10^{-3}$ on resolution of 16,384.

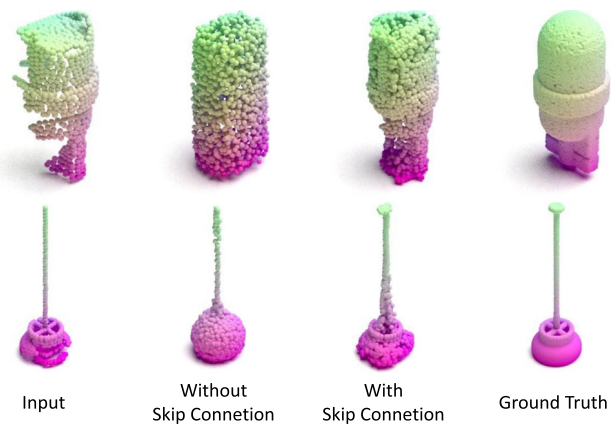

**Fig. 10** Object completion results with and without the influence of the skip-connection

### 7.3 Skip Connection with Feature Transform

One of the key contributions in this paper is the introduction of skip connections with feature transforms on point cloud. Our ablation study in Tables 3 and 2 also includes the numerical advantage of having the skip connection in our architecture, improving the Chamfer distance by $0.65 \times 10^{-4}$ in L1 and $1.27 \times 10^{-4}$ in L2.

In addition to the numerical advantage, we also interpret these values through some examples in Fig. 10 where we reconstruct lamps. Without the skip connection, the model recursively simplifies the given partial scan until it reaches the latent feature. Due to the oversimplification, the output then builds the closest generic shape of the lamp. Contrary to that, with the skip connection, the model preserves the input structure and incorporates the given partial scan into the final reconstruction. In effect, the result is closer to the ground truth.

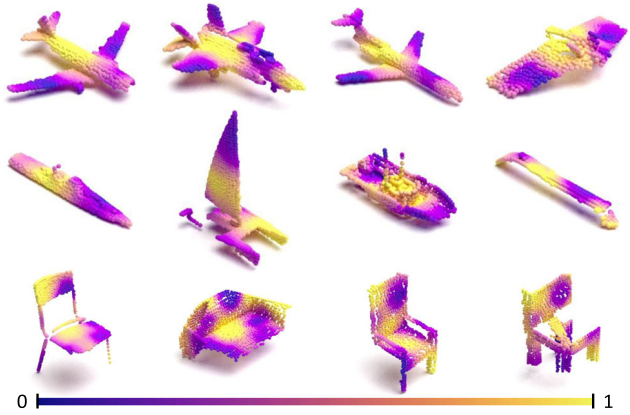

**Fig. 11** Visualization of the first row of **F** on the first SoftPool++ module in our architecture

We also perform an ablation study on the regularization $\mathcal{L}_\mathbf{R}$ of the feature transform **R** in Tables 4, 5, 3 and 2. Compared to our complete framework, the results trained without the skip-connection drops by $0.64 \times 10^{-4}$. However, when trained with the skip-connection but without the regularization of **R**, the results drops by $2.14 \times 10^{-4}$ which is significantly larger. Therefore, it is noteworthy to mention that training with skip-connection but without the regularization performs worse than removing the skip-connection altogether. This clearly shows the advantage of the regularization term on the feature transform.

### 7.4 Activations from the SoftPool++ Features

Given the input point cloud, we explore how SoftPool++ sorts the points on the first feature extraction module in the architecture. For this experiment, we visualize the points based on the value of the first column in **F** which is the result of MLP as shown in Fig. 2. Therefore, Fig. 11 highlights the activations associated to this feature. Noticeably, due to MLP, the points can undergo much more than just a linear transformation of its absolute coordinates.

Continuing our analysis, we move further into examining how the truncation sizes $(N_s, N_r)$ and the output dimension $N_f$ influence the completion. Table 10 summarizes this evaluation on ShapeNet (Chang et al. 2015) as we vary these values on the second SoftPool++ module in our architecture. As described in Table 1, since our SoftPool++ feature is fixed with 256 rows, we then set $N_r \times N_s = 256$. Note that the next ablation study focuses on changing the number of rows by

**Table 10** Influence of $N_f$ and $(N_s, N_r)$ on the L2 Chamfer distance (multiplied by $10^3$), evaluated on the output resolution of 2048 points

| $N_s$ | $N_r$ | $N_f$ | | | | |
|---|---|---|---|---|---|---|
| | | 32 | 64 | 128 | 256 | 512 |
| 1 | 256 | 17.33 | 16.50 | 13.45 | 11.24 | 10.39 |
| 2 | 128 | 17.28 | 17.06 | 15.22 | 14.62 | 13.10 |
| 4 | 64 | 16.32 | 15.66 | 13.65 | 11.29 | 11.31 |
| 8 | 32 | 17.55 | 11.18 | 10.27 | 9.59 | **9.58** |
| 16 | 16 | 17.97 | 11.26 | 10.21 | 9.60 | 9.59 |
| 32 | 8 | 17.31 | 11.89 | 10.32 | 9.59 | **9.58** |

Bold indicates the best performance achieved in certain column

**Table 11** Influence of $N_s$ and $N_r$ on the L2 Chamfer distance (multiplied by $10^3$), evaluated on the output resolution of 2048 points

| $N_s$ | $N_r$ | | | | | |
|---|---|---|---|---|---|---|
| | 8 | 16 | 32 | 64 | 128 | 256 |
| 1 | – | – | 14.99 | 14.82 | 14.64 | 11.24 |
| 2 | – | 14.99 | 14.97 | 14.73 | 14.62 | 11.26 |
| 4 | 14.79 | 12.85 | 12.27 | 11.29 | 11.08 | 9.91 |
| 8 | 11.56 | 10.62 | **9.59** | **9.59** | 9.62 | 9.64 |
| 16 | 10.07 | **9.59** | 9.62 | 9.62 | 9.61 | 9.63 |
| 32 | 9.60 | 9.61 | 9.61 | 9.62 | 9.61 | 9.61 |

Here, $N_f$ is set to 256
Bold indicates the best performance achieved in certain column independently setting $N_r$ and $N_s$. For the ease of training and evaluation for all $(N_s, N_r)$ and $N_f$, we do not apply discriminative training $\mathcal{D}$ for Table 10. The table indicates that we reach the minimum Chamfer distance as soon as $N_s$ reaches 8, $N_r$ reaches 32 and $N_f$ reaches 256. After then, only small improvements of around $0.01 \times 10^{-3}$ are attained. Therefore, we select $N_s = 8$, $N_r = 32$ and $N_f = 256$ so that there are less parameters in the model to train which consequently lead to less memory footprint.

The next ablation study alleviates the constraint of having a fixed latent feature dimension where we set $N_r \times N_s = 256$ in Table 10. In Table 11, we consider different values of $N_r$ and $N_s$ while setting $N_f$ to 256, where we observe that the error plateaus when $N_s$ is 8 and $N_r$ is 32. Note that these values matches the optimum values from Table 10 and validates the advantage of truncation.

Considering the numerical advantages of $N_s$, we also explore it visually while keeping $N_f$ and $N_r$ constant to 256 and 32, respectively. Similar to Fig. 11, Fig. 12 plots the points from the input point cloud that are truncated by $N_s$. By increasing $N_s$ from 4 to 16, the resulting feature also increases the amount of structures from the plane. For instance, the wings become more and more visible on the figure. This then raises the question of how much information from the partial scans does the network need to reconstruct the object. Evidently, this question is answered by our ablation study in Table 10 where we found the optimum value of $N_s$, i.e. 8. Comparisons in Fig. 12 shows that larger values of $N_s$ does not further add the points on the body of the plane which is a common part for plane category.

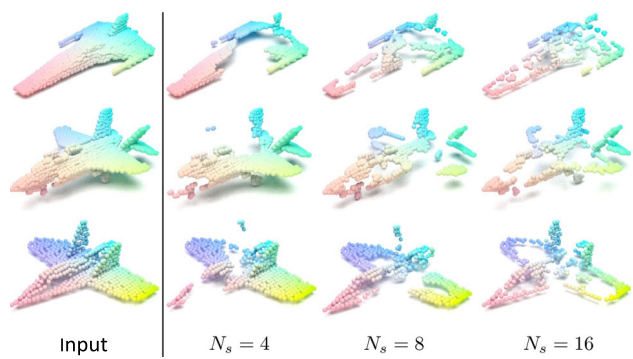

**Fig. 12** Visualization of the truncated points on the input point cloud with different values of $N_s$

## 8 Conclusion

We propose a novel feature extraction technique called *SoftPool++* that directly processes the point cloud. Compared to the counterpart that heavily relies on the max-pooling operation in PointNet (Qi et al. 2017a), our feature extraction method captures a higher amount of data from the input point cloud by alleviating the limitation of taking only the maximum while also establishing the relation between different points through our regional convolutions.

Structuring multiple SoftPool++ in an encoder–decoder structure, this paper becomes the first to propose a point-wise skip connection with feature transformation. Considering that the given point cloud is continuously downsampled in the encoder, the main advantage of such connection is the capacity to incorporate the input data into the decoder. This then overcomes the loss of information in the encoder.

Examining our contributions on 3D object completion, we discovered that we perform the state-of-the-art especially on high-resolution reconstructions. We also visually demonstrate our advantage and concluded that our reconstructions are sharper, i.e. with less noise in our point cloud; and, captures the finer details, i.e. without over-smoothing different parts of the object.

**Funding** Open Access funding enabled and organized by Projekt DEAL.